\documentclass[conference]{IEEEtran}
\usepackage[hidelinks]{hyperref}
\usepackage{subcaption}

\IEEEoverridecommandlockouts
\usepackage{cite}
\usepackage{amsmath,amssymb,amsfonts}
\usepackage{algorithmic}
\usepackage{graphicx}
\usepackage{soul}
\usepackage{stfloats}
\usepackage{textcomp}
\usepackage{xcolor}
\usepackage{subcaption}
\def\BibTeX{{\rm B\kern-.05em{\sc i\kern-.025em b}\kern-.08em
    T\kern-.1667em\lower.7ex\hbox{E}\kern-.125emX}}
\begin{document}

\title{H-Infinity Filter Enhanced CNN-LSTM for
Arrhythmia Detection from Heart Sound
Recordings\\
}
\vspace{-2in}
\author{\IEEEauthorblockN{Rohith Shinoj Kumar, Rushdeep Dinda, Aditya Tyagi, Annappa B., Naveen Kumar M. R. }
\IEEEauthorblockA{\textit{Department of Computer Science and Engineering} \\
\textit{National Institute of Technology Karnataka, Surathkal, Karnataka, India 575025}\\
\textit{rohith.211cs245@nitk.edu.in, rdinda.211cs246@nitk.edu.in, adityatyagi.211cs101@nitk.edu.in,}\\
\textit{annappa@ieee.org, naveenkumarmr.227cs003@nitk.edu.in}} 
\thanks {This is a preprint of a paper to appear at the 15th IEEE International Conference on Systems Engineering and Technology (ICSET 2025).}
}

\maketitle

\begin{abstract}
Early detection of heart arrhythmia can prevent severe future complications in cardiac patients. While manual diagnosis still remains the clinical standard, it relies heavily on visual interpretation and is inherently subjective. In recent years, deep learning has emerged as a powerful tool to automate arrhythmia detection, offering improved accuracy, consistency, and efficiency. Several variants of convolutional and recurrent neural network architectures have been widely explored to capture spatial and temporal patterns in physiological signals. However, despite these advancements, current models often struggle to generalize well in real-world scenarios, especially when dealing with small or noisy datasets, which are common challenges in biomedical applications. In this paper, a novel CNN-H-Infinity-LSTM architecture is proposed to identify arrhythmic heart signals from heart sound recordings. This architecture introduces trainable parameters inspired by the H-Infinity filter from control theory, enhancing robustness and generalization. Extensive experimentation on the PhysioNet CinC Challenge 2016 dataset, a public benchmark of heart audio recordings, demonstrates that the proposed model achieves stable convergence and outperforms existing benchmarks, with a test accuracy of 99.42\% and an F1 score of 98.85\%.
\end{abstract}


\section{Introduction}
Cardiovascular diseases remain the foremost contributor to global mortality, claiming nearly 18 million lives each year \cite{world_2024}. Furthermore, the
number of deaths due to heart disease has risen faster than that of any other cause
worldwide. Heart arrhythmias are irregularities in the heartbeat caused by disrupted electrical signals, leading to rhythms that are too fast (tachycardia), too slow (bradycardia), or erratic. If not identified early, persistent arrhythmias can weaken the heart muscle over time, making it less effective at pumping blood. This increases the chances of the patient suffering from strokes, heart failure, and cardiac arrest. Atrial Fibrillation (AF), for instance, impacts over 2.3 million individuals in the United States alone \cite{khurshid_choi_weng_wang_trinquart_benjamin_ellinor_lubitz_2018}. Early detection and management are crucial to preventing these complications.

\begin{figure}[]
    \centering
    \includegraphics[width=0.3\textwidth]{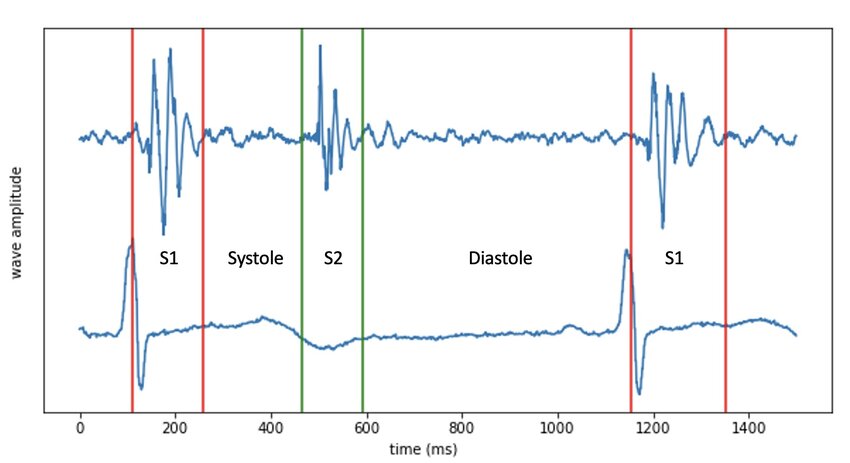}
    \caption{
    Waveform representation of the cardiac cycle phases
    }
    \label{fig:heart_cycle}
\end{figure}

A Phonocardiogram (PCG) is a non-invasive diagnostic tool that graphically represents heart sounds and murmurs, captured using a sensor placed on the chest. Figure \ref{fig:heart_cycle} illustrates the 4 cardiac phases from an ECG and their corresponding interpretation from a standard heart audio sample of a healthy person. 
In this work, the focus is on the classification of cardiac arrhythmias from audio recordings of the cardiac rhythms based on deep learning approaches. 
\begin{figure}[]
    \centering
    \includegraphics[width=0.3\textwidth]{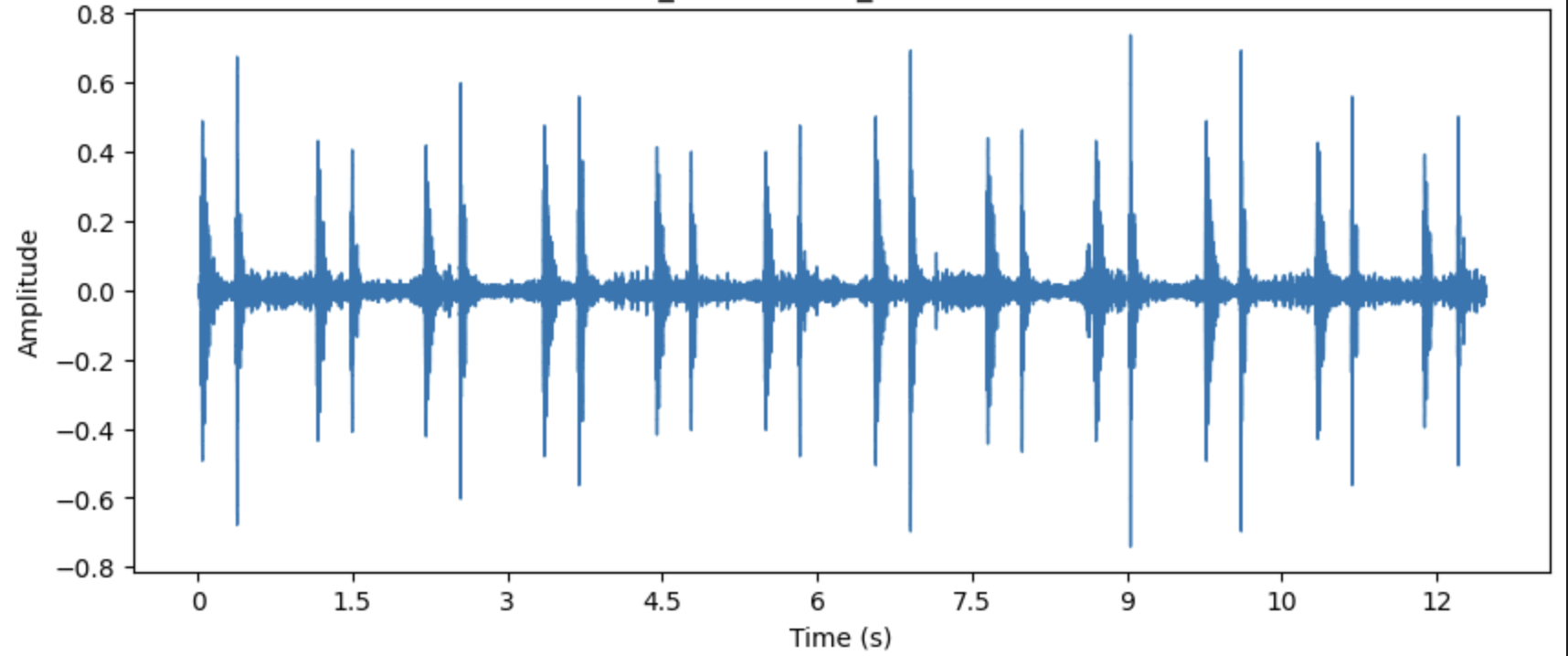}
    
    
    \includegraphics[width=0.3\textwidth]{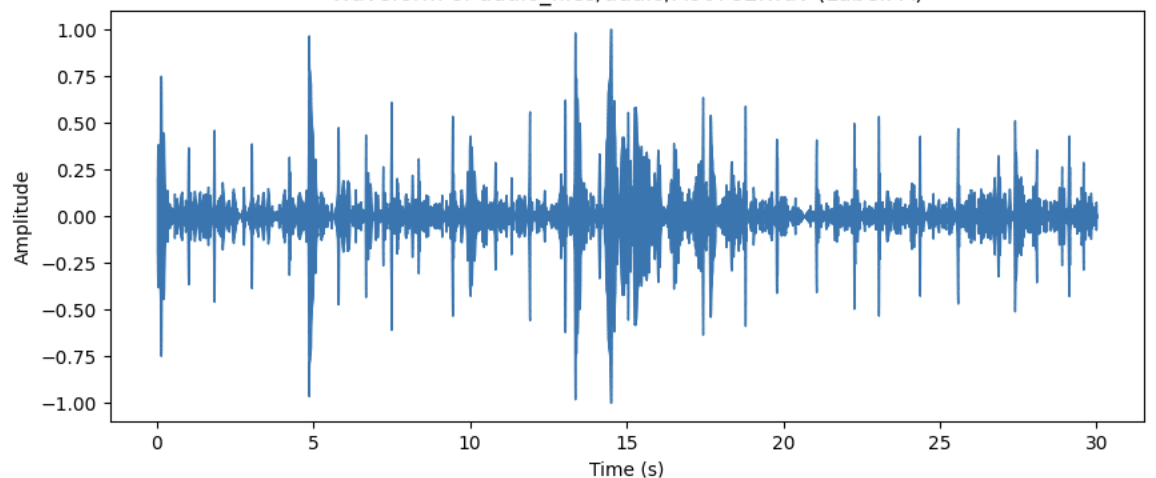}
    
    \caption{
    Healthy (Top) and Arrhythmic (Bottom) Waveforms
    }
    \label{fig:waves}
\end{figure}
  

Heart sound classification has seen substantial advances in recent years, especially through the fusion of spectrograms and deep learning-based techniques. Nilanon et al.~\cite{nilanon} were among the first to propose the use of spectrograms and CNNs for the classification of heart sounds, showing that time-frequency representations can greatly aid in enhanced classification. 

Tsai et al. \cite{murmur} presented a capsule network framework that undertakes the processing of spectrograms by convolutional capsule layers through dynamic routing to enable more hierarchical feature encoding. To tackle class imbalance, Li et al.~\cite{cnn} made use of the weighted loss and further enhanced the input by concatenating the Mel spectrogram with the Mel-Frequency Cepstral Coefficient (MFCC) features to better discriminate normal versus pathological instances. 

Singh-Miller et al.~\cite{spectral} pursued a hybrid approach by applying principal component analysis (PCA) and k-means to extract features that model the activity in different frequency bands, followed by random forest training of these features. Similarly, Vernekar et al.~\cite{vernekar2017novel} extracted a mix of statistical and frequency-domain features, enhanced with Markov-chain analysis, to train a neural network.
Chen et al.~\cite{automatic} devised a CNN-LSTM architecture that takes segmented raw audio as an input to instill both spatial and temporal patterns onto a signal. Deng et al.~\cite{mfcccrnn} proposed a convolutional recurrent neural networks (CRNN) based framework on enhanced MFCC features, experimenting with noise-ridden datasets. $H_\infty$ filters are used in a speech enhancement method presented by \cite{shen1996}, which highlights its potential in eliminating the need for prior knowledge of noise statistics, unlike conventional Wiener and Kalman filtering techniques.

In this work, training and evaluation was performed on recordings from the PhysioNet Computing in Cardiology Challenge (CinC) Arrhythmia Detection dataset \cite{dataset}, a public benchmark of audio recordings annotated for various cardiac conditions. The dataset contains approximately 8500 heart sound recordings at a sampling rate of 2000 Hz, thus containing rich acoustic details necessary for sound diagnostic purposes. However, some of these files are not annotated and such files have been filtered out for the purpose of this work, reducing the effective size of the dataset to roughly 6000 audio files. Figure \ref{fig:waves} illustrates the waveforms of a healthy and arrhythmic sample from the dataset. However, while the dataset is a vast collection of high-quality audio recordings, working with the CinC dataset poses some very serious issues.
\begin{itemize}
\item Pronounced class imbalance: The data show a notable skew, with roughly 87\% (5154 samples) of the recordings classified as normal, and a minority as abnormal heart sounds (771 samples). 

\item Variable recording length: The length of the recordings of the heart sounds varies between a few seconds to over one minute.

\item The existence of noise and artifacts in many real heart sound recordings obscure the underlying cardiac signals.
\end{itemize}
    
    
    
In summary, this work makes two key contributions. Firstly, a novel deep learning architecture, the CNN-$H_\infty$-LSTM, is proposed which replaces the classic forget gate and cell state update equations of an LSTM unit with trainable parameters inspired by the $H_\infty$ filter from control theory. This modification draws on the $H_\infty$ filter's well-established ability to minimize worst-case estimation errors under unknown noise, aiming for better generalization on small and noisy datasets like the one used in this study\cite{HInf}. Secondly, it introduces a training optimization method called Stochastic Adaptive Probe Thresholding (SAPT), coupled with a custom loss function designed to address the issue of class imbalance.
The remaining sections of the paper are organized as follows: Section \ref{section:meth} outlines the proposed methodology, while the experimental setup is illustrated in \ref{section:setup}. The experimental results of the proposed architecture, along with an extensive comparison with prior benchmark models, are presented in Section \ref{section:results}, followed by conclusion and the scope of future work in \ref{section:conc}.

\section{Proposed Methodology}
\label{section:meth}
This section discusses in detail the proposed CNN-$H_\infty$-LSTM architecture trained to identify arrhythmia from a variable-length audio sample of a heart rhythm. This includes noise suppression of the audio sample, generation of Mel spectrograms from the transformed audio waveform and the subsequent training of the proposed architecture with SAPT and custom penalty loss function.

\subsection{Pre-processing for Noise Suppression}

The PhysioNet CinC Challenge Dataset contains audio recordings of heart rhythm that were collected from various clinical and non-clinical settings, and are inherently noisy. These noise components, especially high-frequency artifacts, can drastically impair the performance of arrhythmia detection models.
In this work, a dual-stage pre-processing pipeline for noise suppression has been used. Initially, a Discrete Wavelet Transform is applied to the raw signals to extract multi-scale time-frequency features that capture sustained rhythm patterns. The Discrete Wavelet Transform of a signal \textit{x(t)} is represented by Equation (1).

\begin{equation}
W(j,k) = \int_{-\infty}^{\infty} x(t) \, \psi_{j,k}(t) \, dt
\end{equation}

where \( \psi_{j,k}(t) \) are the scaled and translated versions of the original function\cite{DWT}. The Daubechies 4 (db4) mother waveform was used for this purpose, which is commonly used for signal processing tasks due to its good time-frequency localization properties.

Further, an Infinite Impulse Response (IIR) filter was applied to these transformed signals to smooth the wavelet-processed heart audio and suppress high-frequency noise\cite{IIR}. 



In our implementation, we have used a 5th-order Butterworth low-pass IIR filter with a cutoff frequency of 500 Hz. To standardize the input size and generalize over variable-length audios, we have segmented each heart audio sample into fixed-length 5-second clips. This specific clip duration was chosen to align with the methodology and positive results reported by{\cite{murmur}}.
\begin{figure}[]
    \centering
    \includegraphics[width=0.4\textwidth]{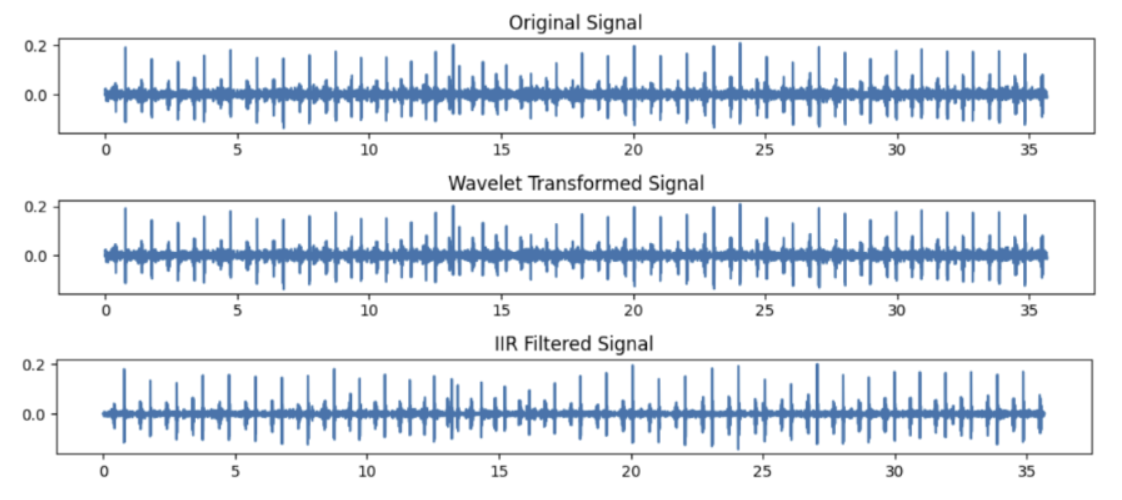}
    \caption{
    Effect of wavelet and IIR Filter on the original signal
    }
    \label{fig:filters}
\end{figure}
\begin{figure}[]
    \centering
    \begin{minipage}[b]{0.24\textwidth}
        \centering
        \includegraphics[width=\textwidth]{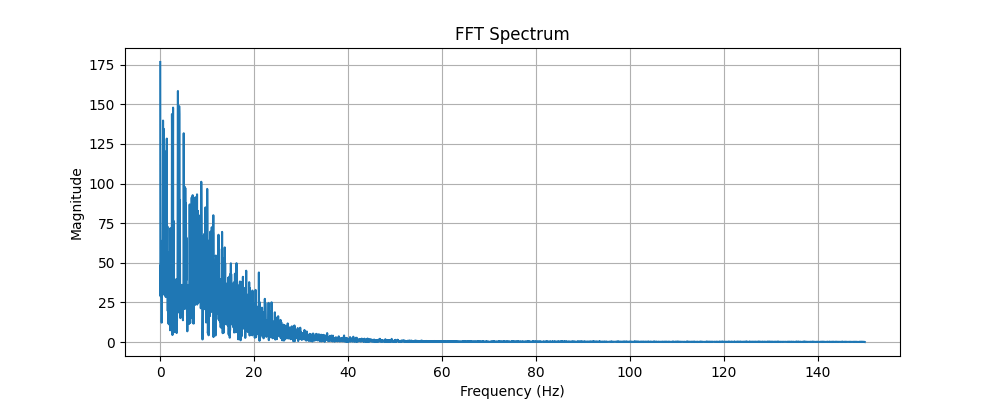}
        \subcaption{Normal signal}
    \end{minipage}
    \hfill
    \begin{minipage}[b]{0.24\textwidth}
        \centering
        \includegraphics[width=\textwidth]{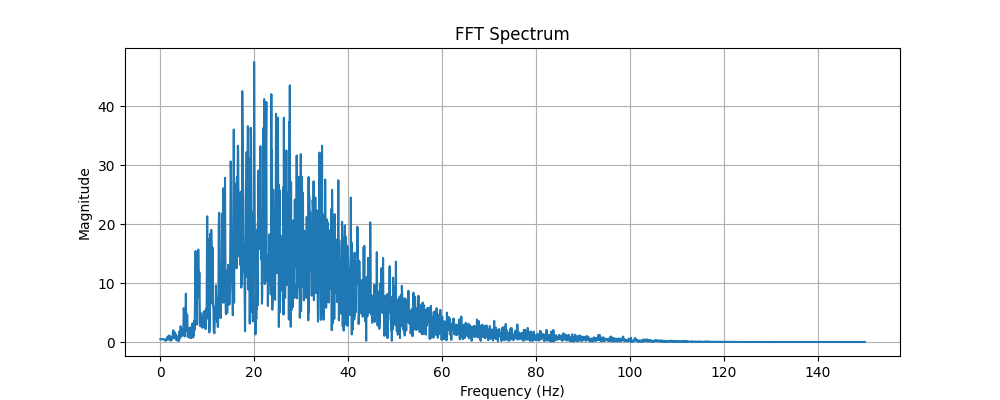}
        \subcaption{Filtered signal}
    \end{minipage}
    \caption{
    Comparison of FFTs of the input and filtered signal.
    }
    \label{fig:FFT}
\end{figure}

Figure \ref{fig:filters} illustrates the incremental effect of the pre-processing pipeline in denoising the raw audio input signal. The final signal retains key rhythmic characteristics and is a cleaner input to the deep learning model. Marked variation is observed in the Fast-Fourier-Transforms of the input and transformed signals as shown in Figure \ref{fig:FFT}

\subsection{Conversion to Mel Spectrogram}
Initial experiments conducted on audio-specific models taking the raw audio files as input did not perform well (discussed further in Section \ref{section:results}). This is attributed to the failure of audio models to recognise the spatial features of the heart rhythm, prompting an exploration of CNN-based image processing models with Mel spectrogram inputs. Mel spectrograms are compact and perceptually relevant descriptions of audio signals. Conversion into Mel spectrogram is ideal for classification tasks because of reduction in dimensionality compared to raw audio
waveforms\cite{mel}. A Mel spectrogram condenses information into fewer, more
meaningful bands instead of working with high-resolution frequency spectrums. Further, Mel spectrograms make the models more robust to variations in background noises in the heart rhythm from the equipment\cite{mel_heart}.

\begin{figure}[]
    \centering
    \begin{minipage}[b]{0.24\textwidth}
        \centering
        \includegraphics[width=\textwidth]{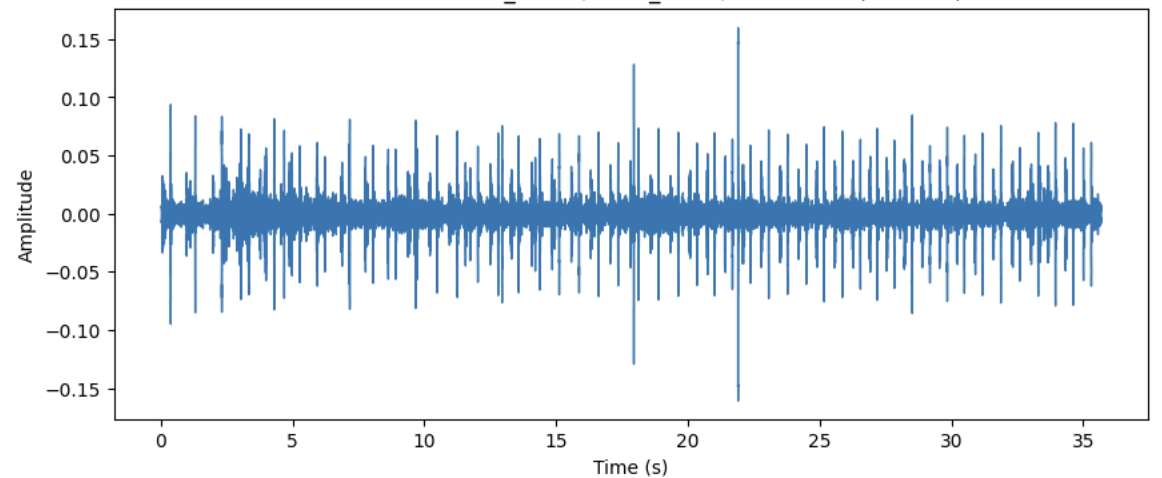}
    \end{minipage}
    \hfill
    \begin{minipage}[b]{0.24\textwidth}
        \centering
        \includegraphics[width=\textwidth]{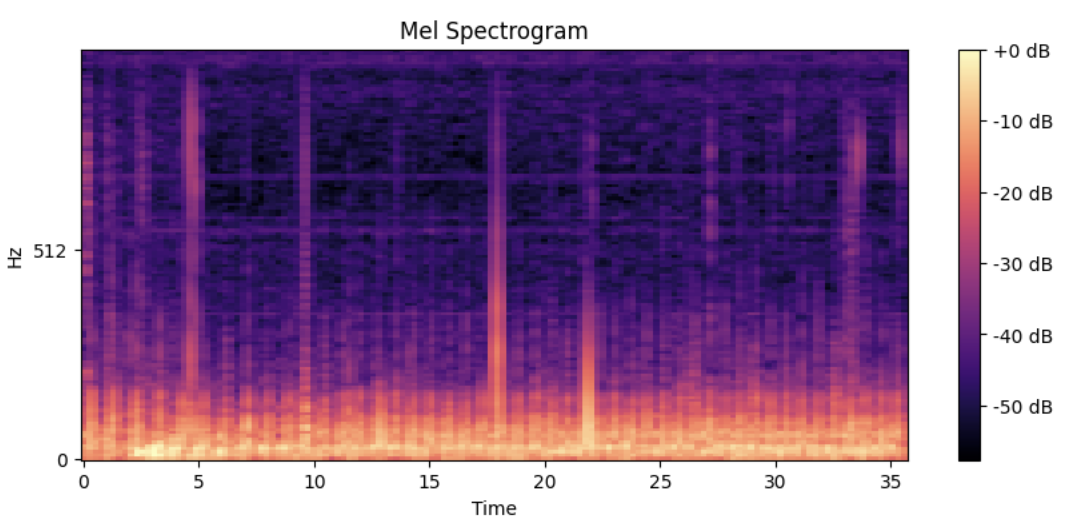}
    \end{minipage}
    \caption{
    Conversion to Mel Spectrogram
    }
    \label{fig:mel}
\end{figure}

Figure \ref{fig:mel} illustrates the transformation of an audio waveform into its Mel spectrogram
representation. The left panel shows the raw audio waveform, displaying amplitude
variations over time. The right panel presents the corresponding Mel spectrogram that is provided as input to the model.

\subsection{Proposed Model}

The input data used for the model in this study, Mel spectrograms, have evident spatial and temporal properties. The proposed architecture introduces a novel method of exploiting the memory retention capabilities of an LSTM network by modifying the internal gate mechanism of the LSTM block. In particular, the default forget gate of the LSTM is substituted with a parameterized $H_\infty$ filter \cite{luan2009} to develop a new recurrent unit that is referred to as the $H_\infty$ LSTM cell. The $H_\infty$ filter excels in minimizing the worst-case estimation error and is particularly useful when underlying noise is non-Gaussian or of unknown nature\cite{noise}. This extension thus allows for better adaptive memory forgetting control for small and noisy datasets in comparison to the traditional forget gate.

The model architecture begins with an input layer designed to receive Mel spectrograms of size ($N_{\text{mels}}$, \(T\), 1)
 where \( N_{\text{mels}} \) represents number of Mel filterbanks and \( T \) represents time steps. This input is passed through a series of convolutional blocks. Each convolutional block comprises two \( 3 \times 3 \) convolutional layers with "same" padding to preserve the dimensionality of the input and followed by a Batch Normalization layer. Batch normalization normalizes across a mini-batch of activations, hence stabilizing and accelerating the training process \cite{bn}. Then, at the end of each convolutional block, a \( 2 \times 2 \) MaxPooling layer reduces the spatial dimensions, thus effectively summarizing and focusing the network's attention on the most important features.

The resulting feature maps are then reshaped along the time axis and passed into the novel $H_\infty$-LSTM module. In this module, each LSTM cell works similarly to a standard LSTM when computing the input and output gates. However, instead of using the usual forget gate, a learnable filter coefficient $\lambda_h$ is introduced, which comes from integrating the $H_\infty$ filter into the LSTM’s gate design. This coefficient helps the model decide how much weight to give to the previous memory state versus the new input, allowing it to learn a more robust, data-driven forgetting mechanism.

The $H_\infty$ filter is designed to provide a guaranteed bound on estimation error even under unknown-but-bounded disturbances and model inaccuracies, making it well-suited for environments with unpredictable or non-stationary noise. Equations 2-5 describe the input gate (\(i_t\)), forget gate (\(f_t\)), output gate (\(o_t\)), and candidate cell state (\(\tilde{c}_t\)) respectively for each block of a standard LSTM.
\vspace{-0.3em}
\begin{equation}
i_t = \sigma(W_i x_t + U_i h_{t-1} + b_i)
\label{eq:input_gate}
\end{equation}
\vspace{-1em}
\begin{equation}
f_t = \sigma(W_f x_t + U_f h_{t-1} + b_f)
\label{eq:forget_gate}
\end{equation}
\vspace{-1em}
\begin{equation}
o_t = \sigma(W_o x_t + U_o h_{t-1} + b_o)
\label{eq:output_gate}
\end{equation}
\vspace{-1em}
\begin{equation}
\tilde{c}_t = \tanh(W_c x_t + U_c h_{t-1} + b_c)
\label{eq:candidate_cell}
\end{equation}
\vspace{-1em}

where $x_t$ is the input vector at time $t$, $h_{t-1}$ is the hidden state from time step $t-1$, $W_i, W_f, W_o, W_c$ represent the weights of the input matrices, $U_i, U_f, U_o, U_c$ represent weights of the recurrent matrices, $b_i, b_f, b_o, b_c$ are bias terms, $\sigma$ is the sigmoid activation function, and $\tanh$ is the hyperbolic tangent activation function \cite{hochreiter1997}.
In the proposed architecture, the input and output gate logic is retained, but the forget gate of a standard LSTM is replaced with a mechanism inspired by the $H_\infty$ filtering approach, creating a robustness coefficient $\lambda_h$ that is derived by passing a trainable parameter $K_\text{filter}$ through a sigmoid activation:
\begin{equation}
\lambda_h = \sigma(K_\text{filter})
\end{equation}
This coefficient dynamically controls the trade-off between retaining past memory $c_{t-1}$ and incorporating new information $i_t \tilde{c}_t$, leading to a modified cell state update:
\begin{equation}
c_t = (1 - \lambda_h) c_{t-1} + \lambda_h i_t \tilde{c}_t
\end{equation}
This structure is mathematically same as the cell state update from the forget gate in a standard LSTM:
\begin{equation}
c_t = f_t c_{t-1} + i_t \tilde{c}_t
\end{equation}
except that instead of computing the forget gate $f_t$ as a function of the current input and hidden state, the model directly learns a fixed robustness coefficient $\lambda_h$.
\begin{figure}[]
    \centering
    \includegraphics[width=0.33\textwidth]{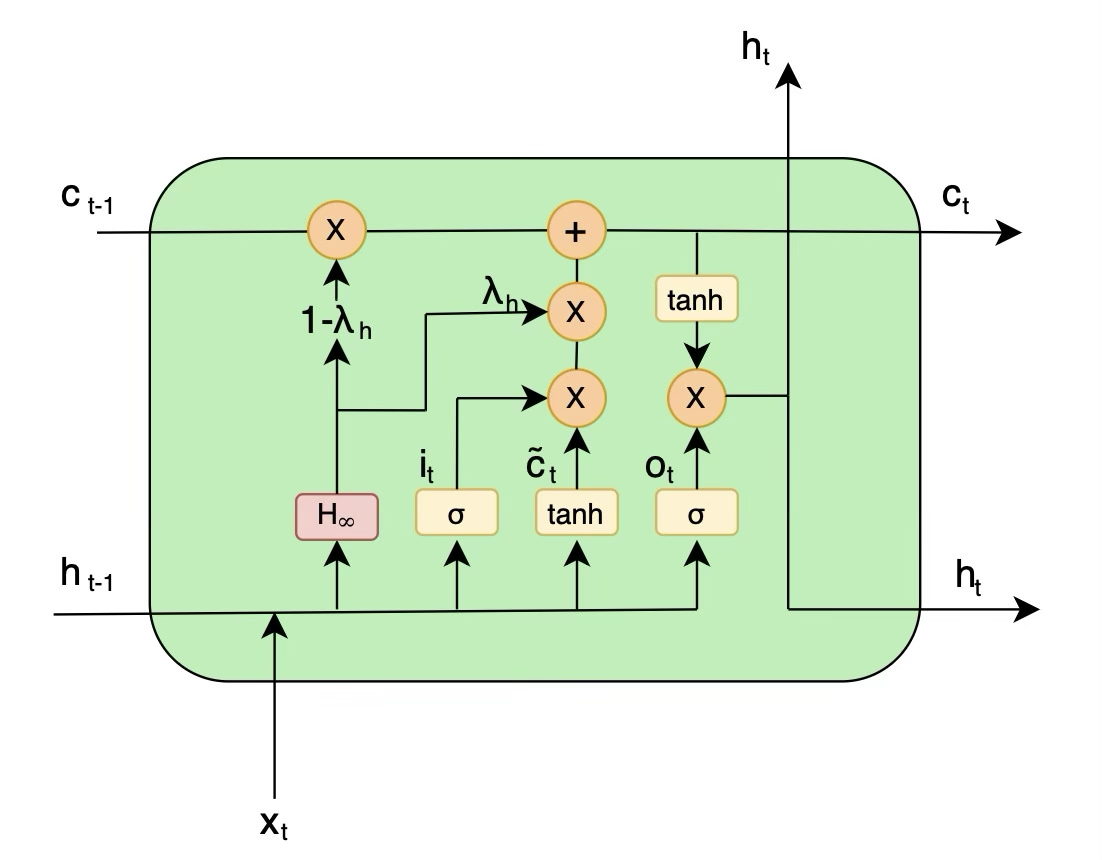}
    \caption{
Structure of a $H_\infty$-LSTM Cell unit
    }
    \label{fig:hlstm}
\end{figure}

\begin{figure}[]
    \centering
    \includegraphics[width=0.47\textwidth]{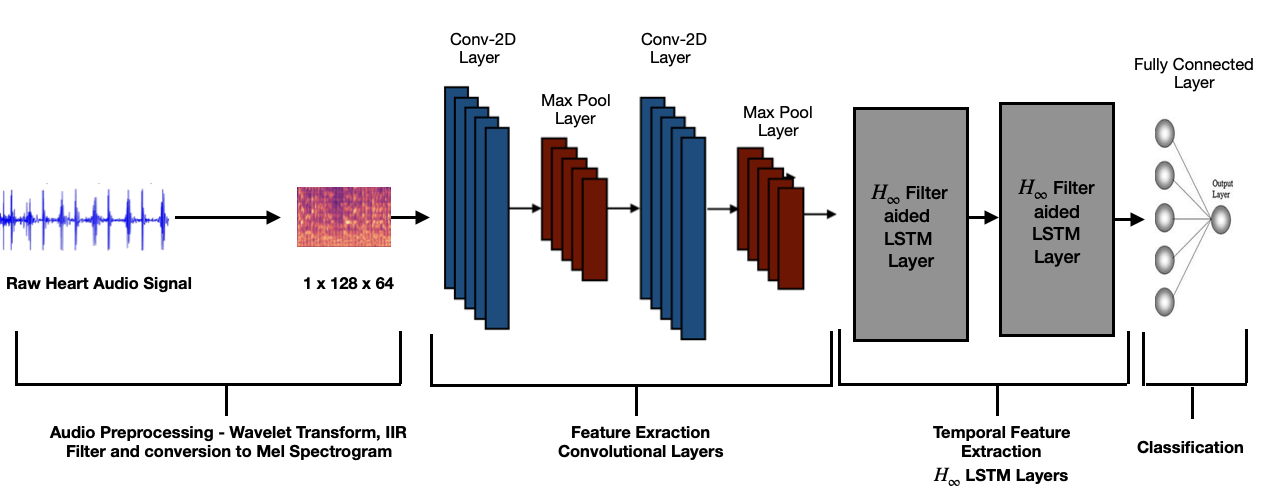}
    \caption{
Proposed Methodology pipeline
    }
    \label{fig:hcnnlstm}
\end{figure}

Figure \ref{fig:hlstm} illustrates the structure of a $H_\infty$-LSTM Cell unit. In contrast to a regular LSTM unit, the $H_\infty$ filter is used in place of the traditional forget gate and is used to control the cell state update across training. The complete proposed model architecture and training pipeline is demonstrated in Fig \ref{fig:hcnnlstm}. 

\subsection{Training Methodology}
\subsubsection{Penalty Weighted Loss (PWL)}
The core idea behind Penalty Weighted Loss (PWL) is to dynamically adjust the contribution of each sample to the loss based on the model’s misclassification behavior, particularly focusing on false negatives and false positives, which are critical in medical diagnosis\cite{BCE}.
Given a batch of size $B$, let the predicted probability vector be $\hat{\mathbf{y}} = (\hat{y}^1, \hat{y}^2, \dots, \hat{y}^B)$ and the ground truth labels be $\mathbf{y} = (y^1, y^2, \dots, y^B)$, where $y^i \in \{0, 1\}$. Define a decision threshold $\delta \in [0, 1]$.

\paragraph{False Negative Index (FNI):}
\begin{equation}
\text{FNI}(\delta) = \sum_{i=1}^{B} \mathbb{I}(\hat{y}_i \leq \delta \land y_i = 1)
\end{equation}

\paragraph{False Positive Index (FPI):}
\begin{equation}
\text{FPI}(\delta) = \sum_{i=1}^{B} \mathbb{I}(\hat{y}_i \geq \delta \land y_i = 0)
\end{equation}
Here, $\mathbb{I}(\cdot)$ denotes the indicator function, that evaluates to 1 when the specified condition holds true, and 0 otherwise.

To penalize misclassifications adaptively, the penalty term is defined as:
\begin{equation}
\mathcal{R}_{\text{penalty}}(\delta) = 1 + \alpha \cdot \text{FNI}(\delta) + (1 - \alpha) \cdot \text{FPI}(\delta)
\end{equation}

where $\alpha \in (0, 1)$ balances the emphasis between false negatives and false positives.

The standard Binary Cross-Entropy (BCE) loss over the batch is:
\begin{equation}
\mathcal{L}_{\text{BCE}} = -\frac{1}{B} \sum_{i=1}^{B} \left[ y_i \log(\hat{y}_i) + (1 - y_i) \log(1 - \hat{y}_i) \right]
\end{equation}

The proposed Penalty Weighted Loss (PWL) becomes:
\begin{equation}
\mathcal{L}_{\text{PWL}} = \mathcal{R}_{\text{penalty}}(\delta) \cdot \mathcal{L}_{\text{BCE}}
\end{equation}

\begin{figure}[]
    \centering
    \includegraphics[width=0.4\textwidth]{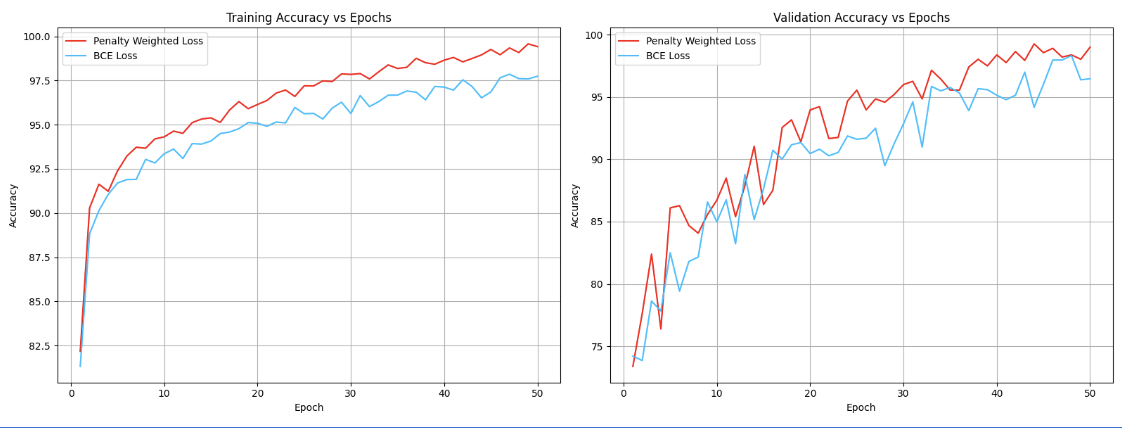}
    \caption{
    Comparison of the performance of CNN-$H_\infty$-LSTM model trained under various loss functions
    }
    \label{fig:losses}
\end{figure}

This increases the loss proportionally to the number of misclassifications, thus compelling the model to focus more on minority class errors, particularly false negatives, which are crucial in clinical settings. Figure \ref{fig:losses} compares the training and validation performance
of the same proposed model when the custom Penalty Weighted Loss
(PWL) is used in comparison to the Binary Cross Entropy Loss described in Equation (14). A marked increase in accuracy for both training and validation is observed using PWL
when other training conditions remain the same. 

\vspace{0.6 em}

\subsubsection{Stochastic Adaptive Probe Thresholding (SAPT)}
While the loss function guides the model to reduce prediction errors, selecting an optimal classification threshold $\tau$ is equally vital, especially in imbalanced datasets. A fixed threshold ($\tau = 0.5$) often biases the model towards the majority class, adversely impacting sensitivity metrics such as recall.

SAPT introduces an adaptive framework to dynamically learn the decision threshold $\tau^*$ during training over the threshold space $[0, 1]$.

The objective is to find a threshold $\tau^*$ that maximizes a task-specific metric $M(\tau)$, such as F1-score, balanced accuracy, or recall.

\begin{enumerate}
    \item Discretization: At a predefined epoch interval $\gamma$, evaluate a set of candidate thresholds $T = \{\tau_j \in [0,1]\}$ 

    \item Evaluation: At each epoch $t$, compute the F1-score $F_t(\tau_j)$ for each threshold $\tau_j$ on the validation set.

    \item Stochastic Optimization: To account for the noise in model predictions, adopt a stochastic approximation approach where thresholds are sampled and updated based on recent performance estimates.

    \item Threshold Update: During each $\gamma$-epoch window, select the threshold $\tau_t$ that maximizes the expected F1-score.



    \item Adaptive Decision Boundary: The selected threshold $\tau^*_t$ is used in the subsequent epoch for classification, allowing the model to adapt to changes in the data distribution or model calibration during training.
\end{enumerate}

\section{Experimental Setup}
\label{section:setup}
To ensure fair comparison, a consistent setup has been followed across all models. Each model was trained for 50 epochs with a fixed learning rate of 0.001 using the Adam optimizer. The dataset was partitioned into an 80:20 training and testing split. An additional validation set, comprising 250 healthy and 250 unhealthy samples, was strictly reserved for hyperparameter tuning. To ensure the integrity and generalization of the model, the testing set was entirely isolated from the training process. Further, no data segments originating from the same patient were assigned to the training and testing partitions simultaneously, thereby preventing data leakage. 

A fine-tuning strategy was adopted, where a CNN-LSTM model was trained on the dataset, and then the weights of the CNN layers were transferred to the CNN-$H_\infty$-LSTM model. The CNN layers were then frozen, and the weights of the $H_\infty$-LSTM layers were trained.
This isolates the adaptation to the layers that matter most for handling time dependencies under noise and imbalance, while leveraging the pre-learned representations for spatial features. 

Further, in the experiments, the Stochastic Adaptive Probe Thresholding (SAPT) hyperparameter \(\gamma\) has been set to 10 based on experimental tuning. This provides the model with sufficient iterations to smooth performance metrics over time as lower values resulted in overly reactive threshold updates, causing erratic behavior in early training epochs. Higher values, on the other hand, slowed the adaptation and reduced its generalization. The Exponential Weighted Moving Average (EWMA) smoothing factor used in this case was 0.3, while the initial classification threshold $\tau$ remains set to the default 0.5. Additionally, the hyperparameter \(\alpha\) in the Penalty Weighted Loss is set to a value of 0.87, which is roughly equal to the class imbalance ratio of the dataset used.

\section{Results}
\label{section:results} 
In this section, a comparative analysis is conducted between the proposed model, audio specific models, vision models and models proposed in heart sound classification literature. Accuracy, F1 score, sensitivity and specificity has been used as metrics to evaluate different models comprehensively.

\subsection{Audio Models}
The first experiments were with end-to-end audio-based deep learning models. The Wav2Vec, HuBERT, and DistilHuBERT architectures were tested on the raw audio recordings. Since these models are fine-tuned for speech tasks, they did not generalize well with heart sound data. The classification results were unsatisfactory, and the models showed poor results in distinguishing murmurs. The results are summarized in Table~\ref{tab:audio_results}.

\begin{table}[!t]
  \centering
  \caption{Performance of pre-trained audio models on heart sound classification}
  \label{tab:audio_results}
  \resizebox{\columnwidth}{!}{%
    \begin{tabular}{lcccc}
      \hline
      \textbf{Model}       & \textbf{F1 Score (\%)} & \textbf{Accuracy (\%)} & \textbf{Sensitivity (\%)} & \textbf{Specificity (\%)} \\
      \hline
      Wave2Vec2 & 69.54 & 68.14 & 63.26 & 63.24 \\
      HuBERT & 71.21 & 71.21 & 71.85 & 70.58 \\
      DistilHuBERT & 74.45 & 78.23 & 72.47 & 76.54 \\
      \hline
    \end{tabular}%
  }
\end{table}

\subsection{Vision Models}
The next experiments tested state of the art image classification models with the Log-Mel Spectrograms of the segmented audio clips. Architectures that were considered included ResNet, MobileNet and Vision Transformers. The ResNet-50 model attained the best results with an F1 score of 89.68\%  and an accuracy of 88.94\%. While the Vision Transformer had promising results in specificity, the architecture requires a lot more data to fully leverage its representational power. Table~\ref{tab:pretrained_results} presents an overview of the obtained results.\\

\begin{table}[!b]
 \centering
 \caption{Performance of pre-trained vision models on heart sound classification}
 \label{tab:pretrained_results}
 \resizebox{\columnwidth}{!}{%
  \begin{tabular}{l c c c c}
   \hline
   \textbf{Model}          & \textbf{F1 Score (\%)} & \textbf{Accuracy (\%)} & \textbf{Sensitivity (\%)} & \textbf{Specificity (\%)} \\
   \hline
   ResNet-50 & 89.68                  & 88.94 & 93.83 & 83.82 \\
   MobileNetV3-Large & 81.36 & 81.27 & 79.01 & 83.82 \\
   Vision Transformer & 84.23 & 95.23 & 89.97 & 96.08 \\
   \hline
  \end{tabular}%
 }
\end{table}
\vspace{-0.7 em}

\subsection{Proposed Model}
\begin{figure}[!b]
  \centering
  \setkeys{Gin}{width=0.48\columnwidth}%
  \subfloat[Training Accuracy]{%
    \includegraphics{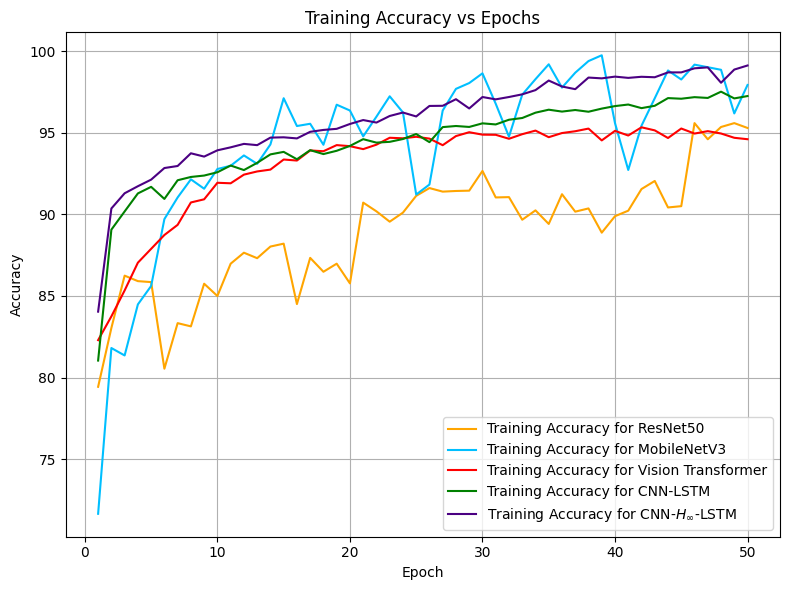}%
    \label{fig:acc}%
  }\hfill
  \subfloat[Validation Accuracy]{%
    \includegraphics{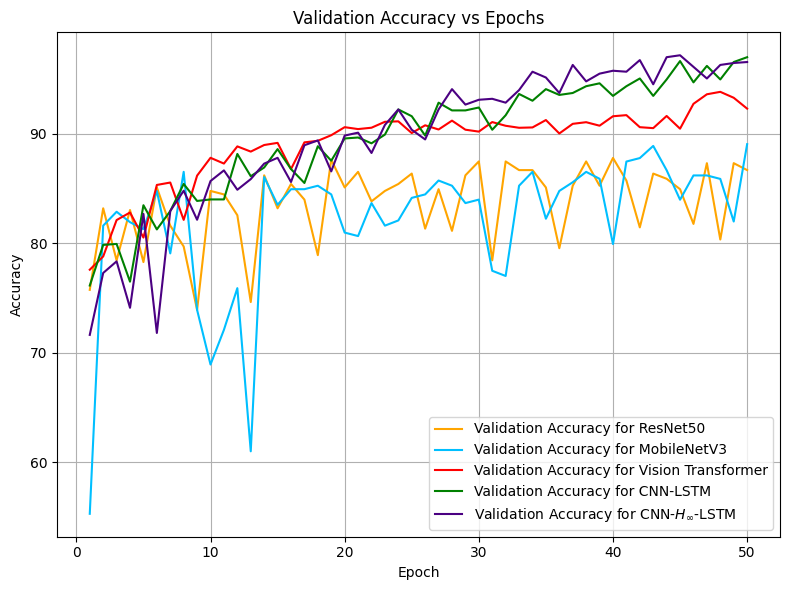}%
    \label{fig:loss}%
  }
  \caption{Accuracy Curves For benchmarked models}
  \label{fig:acc-graph}
\end{figure}

\begin{figure}[!t]
  \centering
  \setkeys{Gin}{width=0.48\columnwidth}%
  \subfloat[Training Loss]{%
    \includegraphics{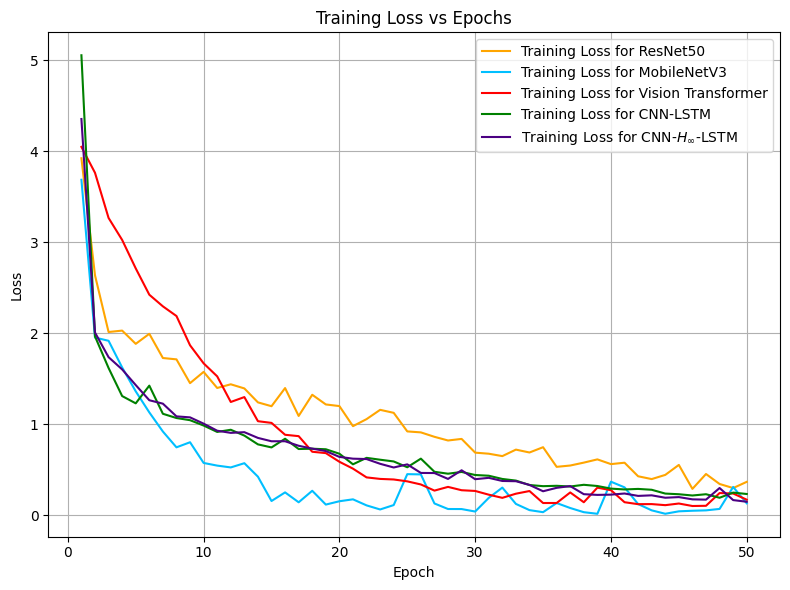}%
    \label{fig:acc}%
  }\hfill
  \subfloat[Validation Loss]{%
    \includegraphics{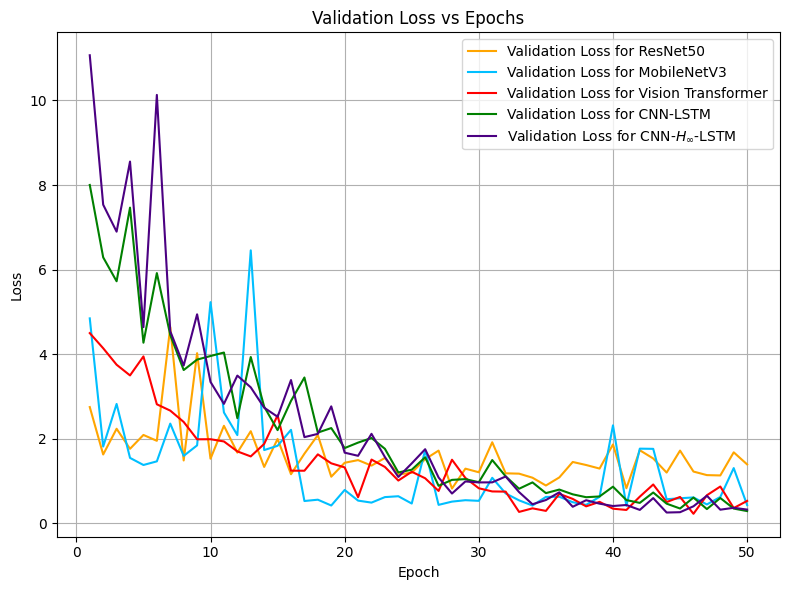}%
    \label{fig:loss}%
  }
  \caption{Loss Curves for benchmarked models}
  \label{fig:loss-graph}
\end{figure}

The proposed design augments the CNN-LSTM framework by adding an $H_\infty$ filter in the LSTM cell. The CNN layers identify spatial features, while the LSTM layers track temporal dependencies.\\
Figs.~\ref{fig:acc-graph}-\ref{fig:loss-graph} show the improvement in training and validation results achieved by the proposed model and previous studies over successive epochs.\\

\vspace{-1em}
\subsection{Performance comparison with existing approaches}
This section compares the proposed model’s performance evaluated on the CinC PhysioNet 2016 Dataset. A quantitative comparison between test performances of proposed model and existing benchmarks is provided in Table~\ref{tab:proposed_results}.\\

\begin{table}[!t]
    \centering
    \caption{Performance comparison between the proposed model and top-performing baseline models}
    \label{tab:proposed_results}
    \resizebox{\columnwidth}{!}{%
    \begin{tabular}{lcccc}
        \hline
        \textbf{Model} & \textbf{F1 (\%)} & \textbf{Acc.\ (\%)} & \textbf{Sens.\ (\%)} & \textbf{Spec.\ (\%)} \\
        \hline
        \textbf{CNN-$H_\infty$-LSTM with SAPT}     & \textbf{98.85} & \textbf{99.42} & \textbf{99.23} & \textbf{99.49} \\
        CNN-LSTM with SAPT                         & 96.19          & 98.16          & 94.69          & 99.29          \\
        ResNet-50                                  & 89.68          & 88.94          & 93.83          & 83.82          \\
        MobileNetV3-Large                          & 81.36          & 81.27          & 79.01          & 83.82          \\
        Vision Transformer                         & 84.23          & 95.23          & 89.97          & 96.08          \\
        Log-Mel VGGNet \cite{cnn}                  & --             & --             & 89.5           & 89.7           \\
        LSTM-CNN \cite{automatic}                  & 91             & 86             & 87             & 82             \\
        Capsule Neural Network \cite{murmur}       & 91             & 90             & 84.87          & --             \\
        CRNN \cite{mfcccrnn}                       & 98.34          & 98.34          & 98.66          & 98.01          \\
        \hline
    \end{tabular}%
    }
\end{table}
\vspace{-1em}
The proposed framework achieves an F1 score of 98.85\% and accuracy of 99.42\%, outperforming the pretrained models and previous studies.
The gains in performance can be attributed to the model's specialized architecture and the new training methodology proposed in this work.\\

\section{Conclusion}
\label{section:conc}
In this paper, the CNN-$H_\infty$-LSTM is proposed as a novel deep learning architecture for automated arrhythmia detection from heart sound recordings. The proposed method includes using mel spectrogram of the heart sound recordings as input, replacing the traditional forget gate of a CNN-LSTM with an adaptive $H_\infty$ filter to improve robustness against noise and variability. While the traditional CNN-LSTM relies on a fixed forget gate that is susceptible to noisy and variable signals, the integration of the $H_\infty$ filter allows the model to achieve greater robustness and dynamic state correction for improved arrhythmia detection, even under heavier noise. To address extreme class imbalance, a custom training strategy, SAPT, has been introduced which improves convergence stability and minority class recall. Evaluated on the PhysioNet 2016 CinC Challenge dataset, the proposed method achieves 98.85\% F1-score, 99.42\% accuracy, 99.23\% sensitivity, and 99.49\% specificity, outperforming prior approaches. The end-to-end design enables real-time deployment on mobile or edge devices, supporting scalable and low-cost cardiac screening. Future work includes integrating the $H_\infty$ filter to more advanced architectures, coupled with centroid-based thresholding and evaluating AutoBalance optimizer for further improvements in class imbalance handling and clinical applicability.

\bibliographystyle{ieeetr}
\bibliography{ref.bib}
\end{document}